\documentclass[10pt,twocolumn,letterpaper]{article}

\usepackage{iccv}
\usepackage{times}
\usepackage{epsfig}
\usepackage{graphicx}
\usepackage{amsmath}
\usepackage{amssymb}
\usepackage{mathrsfs}
\usepackage{algorithm}
\usepackage{algorithmic}
\usepackage{bbding}
\usepackage{pifont}
\usepackage{subfigure}
\usepackage{multirow}
\usepackage{enumitem}
\usepackage{authblk}

\usepackage[breaklinks=true,bookmarks=false]{hyperref}

\iccvfinalcopy 

\def\iccvPaperID{4042} 

\ificcvfinal\fi

\makeatletter
\newcommand{\printfnsymbol}[1]{%
  \textsuperscript{\@fnsymbol{#1}}%
}
\makeatother

\begin{document}

\title{Online Hyper-parameter Learning for Auto-Augmentation Strategy}

\author{Chen Lin\textsuperscript{1}\thanks{equal contribution}, Minghao Guo\textsuperscript{1}\printfnsymbol{1}, Chuming Li\textsuperscript{1}, Xin Yuan\textsuperscript{1}, Wei Wu\textsuperscript{1}, Junjie Yan\textsuperscript{1}, Dahua Lin\textsuperscript{2}, Wanli Ouyang\textsuperscript{3}\\
\textsuperscript{1}SenseTime Group Limited \\
\textsuperscript{2}The Chinese University of Hong Kong \\
\textsuperscript{3}The University of Sydney \\
{\tt\small \{linchen,guominghao,lichuming,yuanxin,wuwei,yanjunjie\}@sensetime.com; dhlin@ie.cuhk.edu.hk; wanli.ouyang@sydney.edu.au}
}

\maketitle
\ificcvfinal\thispagestyle{empty}\fi

\begin{abstract}
Data augmentation is critical to the success of modern
deep learning techniques. In this paper, we propose
Online Hyper-parameter Learning for Auto-Augmentation
(OHL-Auto-Aug), an economical solution that learns the
augmentation policy distribution along with network training.
Unlike previous methods on auto-augmentation that
search augmentation strategies in an offline manner, our
method formulates the augmentation policy as a parameterized
probability distribution, thus allowing its parameters
to be optimized jointly with network parameters. Our proposed
OHL-Auto-Aug eliminates the need of re-training and
dramatically reduces the cost of the overall search process,
while establishes significantly accuracy improvements over
baseline models. On both CIFAR-10 and ImageNet, our
method achieves remarkable on search accuracy, i.e. 60$\times$
faster on CIFAR-10 and 24$\times$ faster on ImageNet, while
maintaining competitive accuracies.
\end{abstract}

\section{Introduction}
Recent years have seen remarkable success of deep
neural networks in various computer vision applications.
Driven by large amounts of labeled data, the performances
of networks have reached an amazing level. A common issue
in learning deep networks is that the training process
is prone to overfitting, with the sharp contrast between the
huge model capacity and the limited dataset. Data augmentation,
which applies randomized modifications to the training
samples, have been shown to be an effective way to mitigate
this problem. Simple image transforms, including random
crop, horizontal flipping, rotation and translation, are
utilized to create label-preserved new training data to promote
the network generalization and accuracy performance. Recent efforts such as \cite{devries2017improved, zhang2017mixup} further develop the transform strategy and boost the model to state-of-the-art accuracy on CIFAR-10 and ImageNet datasets. 

\begin{figure}[tb]
  \begin{center}
    \includegraphics[width=1.0\linewidth]{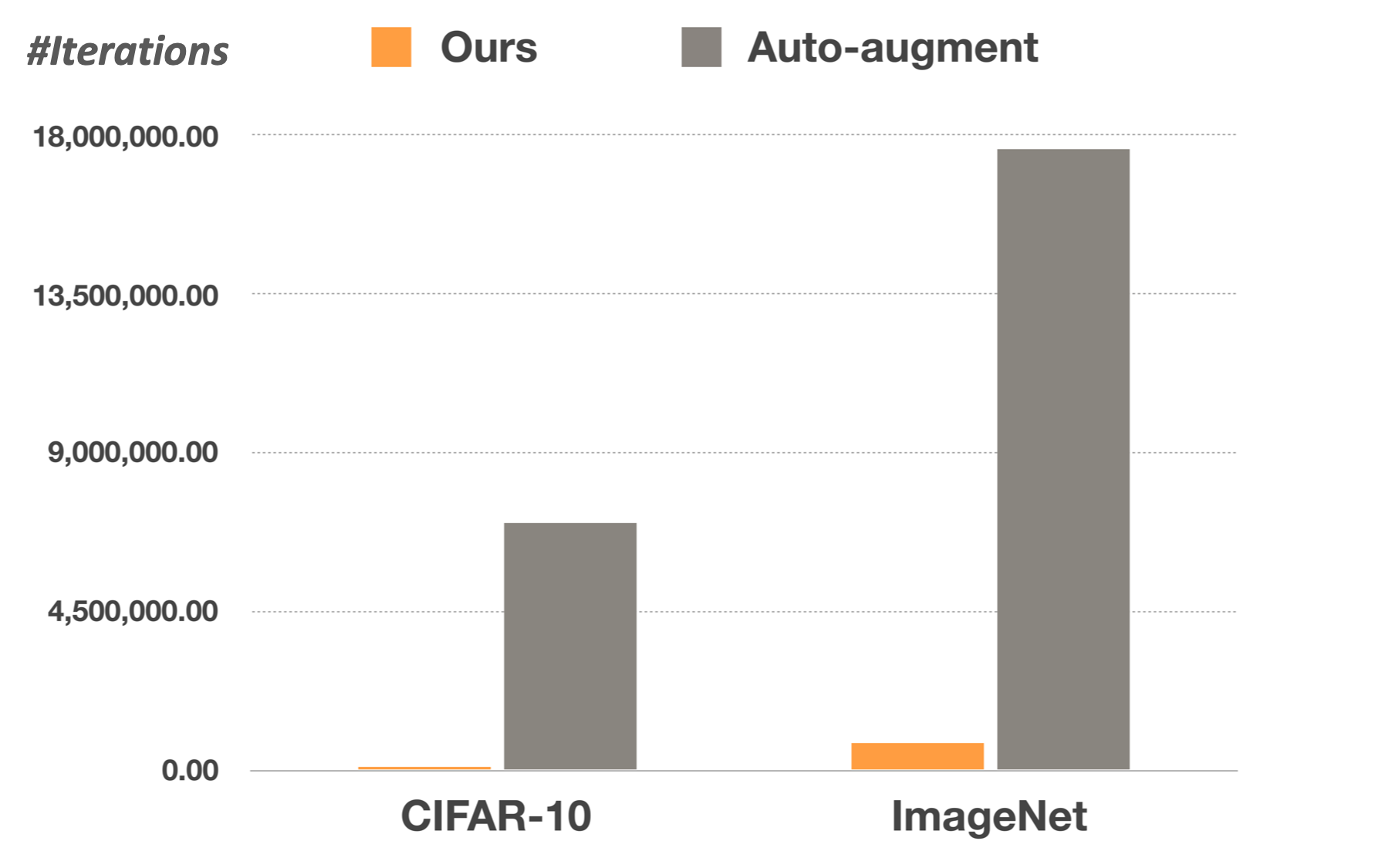}
  \end{center}
  \caption{\small{The search cost of Auto-Augment \cite{cubuk2018autoaugment} and our OHL-Auto-Aug. `$\#Iterations$' denotes the total training iterations with conversion under a same batch size 1024 (more details in Section \ref{SearchCost}). Our proposed OHL-Auto-Aug achieves a remarkable auto-augmentation search efficiency, which is 60$\times$ faster on CIFAR-10 and 24$\times$ faster on ImageNet than Auto-Augment \cite{cubuk2018autoaugment}.}}
  \label{fig-motivation}
\end{figure}

However, it is nontrivial to find a good augmentation policy
for a given task, as it varies significantly across datasets
and tasks. It often requires extensive experience combined
with tremendous efforts to devise a reasonable policy. Instinctively,
automatically finding the process of data augmentation
strategy for a target dataset turns into an alternative.
Cubuk \etal~\ \cite{cubuk2018autoaugment} described a procedure to automatically
search augmentation strategy from data. The search
process involves sampling hundreds and thousands of policies,
each trained on a child model to measure the performance
and further update the augmentation policy distribution
represented by a controller. Despite its promising empirical
performance, the whole process is computationally
expensive and time consuming. Particularly, it takes 15000
policy samples, each trained with 120 epochs. This immense
amount of computing limits its applicability in practical
environment. Thus \cite{cubuk2018autoaugment} subsampled the dataset to $8\%$ and $0.5\%$ for CIFAR-10 and ImageNet respectively. An inherent
cause of this computational bottleneck is the fact that
augmentation strategy is searched in an offline manner.

In this paper, we present a new approach that considers
auto-augmentation as a hyper-parameter optimization
problem and drastically improves the efficiency of the autoaugmentation
strategy. Specifically, we formulate augmentation policy as a probability distribution. The parameters
of the distribution are regarded as hyper-parameters. We
further propose a bilevel framework which allows the distribution
parameters to be optimized along with network
training. In this bilevel setting, the inner objective is the
minimization of vanilla train loss w.r.t network parameters,
while the outer objective is the maximization of validation
accuracy w.r.t augmentation policy distribution parameters
utilizing REINFORCE \cite{williams1992simple}. These two objectives are optimized
simultaneously, so that the augmentation distribution
parameters are tuned alongside the network parameters
in an online manner. As the whole process eliminates
the need of retraining, the computational cost is exceedingly
reduced. Our proposed OHL-Auto-Aug remarkably
improves the search efficiency while achieving significant
accuracy improvements over baseline models. This ensures
our method can be easily performed on large scale dataset
including CIFAR-10 and ImageNet without any data reduction.

Our main contribution can be summaried as follows: \textbf{1)} We propose an online hyper-parameter learning approach
for auto-augmentation strategy which treats each augmentation
policy as a parameterized probability distribution. \textbf{2)} We introduce a bilevel framework to train the distribution
parameters along with the network parameters, thus
eliminating the need of re-training during search. \textbf{3)} Our
proposed OHL-Auto-Aug dramatically improves the efficiency
while achieving significant performance improvements.
Compared to the previous state-of-the-art auto-augmentation
approach \cite{cubuk2018autoaugment}, our OHL-Auto-Aug achieves 60$\times$ faster on CIFAR-10 and 24$\times$ faster on ImageNet with comparable accuracies.

\section{Related Work}

\begin{figure*}[tb]
\centerline{\includegraphics[width=1.9\columnwidth]{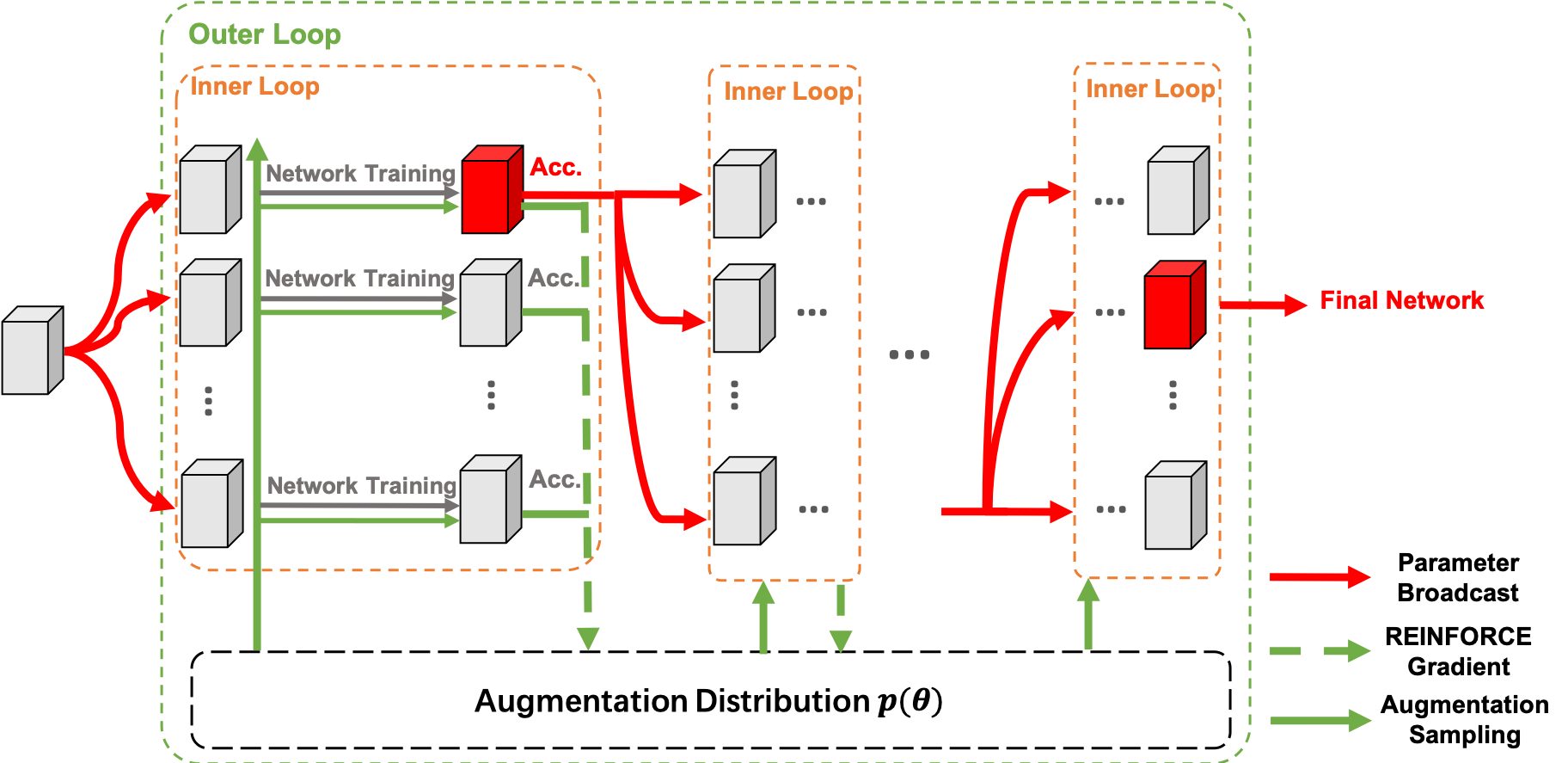}}
\caption{The framework of our OHL-Auto-Aug. We formulate each augmentation policy as a parameterized probability distribution, whose parameters are regarded as hyper-parameters. We propose a bilevel framework which allows the distribution parameters to be optimized simultaneously with the network parameters. In the inner loop, the network parameters are trained using standard SGD with augmentation sampling. In the outer loop, augmentation policy distribution parameters are trained using REINFORCE gradients with trajectory samples. At each time step, the network parameters with the highest accuracy are broadcasted to other trajectory samples.}
\label{fig:framework}
\end{figure*}

\subsection{Data Augmentation}
Data augmentation lies at the heart of all successful applications of deep neural networks. In order to improve the network generalization, substantial domain knowledge is leveraged to design suitable data transformations. \cite{lecun1998gradient} applied various affine transforms, including horizontal and vertical translation, squeezing, scaling, and horizontal shearing to improve their model’s accuracy. \cite{krizhevsky2012imagenet} exploited principal component analysis on ImageNet to randomly adjust colour and intensity values. \cite{wu2015deep} further utilized a wide range of colour casting, vignetting, rotation, and lens distortion to train Deep Image on ImageNet. 

There also exists attempts to learn data augmentations strategy, including Smart Augmentation \cite{lemley2017smart}, which proposed a network that automatically generates augmented data by merging two or more samples from the same class, and \cite{tran2017bayesian} which used a Bayesian approach to generate data based on the distribution learned from the training set. Generative adversarial networks have also been used for the purpose of generating additional data \cite{perez2017effectiveness,zhu2017data,antoniou2017data}.  The work most closely related to our proposed method is \cite{cubuk2018autoaugment}, which formulated the auto-augmentation search as a discrete search problem and exploited a reinforcement learning framework to search the policy consisting of possible augmentation operations. Our proposed method optimizes the distribution on the discrete augmentation operations but is much more computationally economical benefiting from the proposed hyper-parameter optimization formulation.

\subsection{Hyper-parameter Optimization}
Hyper-parameter optimization was historically accomplished by selecting several values for each hyper-parameter, computing the Cartesian product of all these values, then running a full training for each set of values. \cite{bergstra2012random} showed that performing a random search was more efficient than a grid search by avoiding excessive training with the hyper-parameters that were set to a poor value. \cite{bousquet2017critical} later refined this random search process by using quasi random search. \cite{snoek2012practical} further utilized Gaussian processes to model the validation error as a function of the hyper-parameters. Each training further refines this function to minimize the number of sets of hyper-parameters to try. All these methods are “black-box” methods since they assume no knowledge about the internal training procedure to optimize hyper-parameters. Specifically, they do not have access to the gradient of the validation loss w.r.t. the hyper-parameters. To address this issue, \cite{maclaurin2015gradient, franceschi2017forward} explicitly utilized the parameter learning process to obtain such a gradient by proposing a Lagrangian formulation associated with the parameter optimization dynamics. \cite{maclaurin2015gradient} used a reverse-mode differentiation approach, where the dynamics corresponds to stochastic gradient descent with momentum. The theoretical evidence of the hyper-parameter optimization in our auto-augmentation is \cite{franceschi2017forward}, where the hyper-parameters are changed after a certain number of parameter updates in a forward manner. The forward-mode procedure is suitable for the auto-augmentation search process with a drastically reduced cost.

\subsection{Auto Machine Learning and Neural Architecture Search}
Auto Machine Learning (AutoML) aims to free human practitioners and researchers from these menial tasks. Recently many advances focus on automatically searching neural network architectures. One of the first attempts \cite{zoph2017learning,zoph2016neural} was utilizing reinforcement learning to train a controller that represents a policy to generate a sequence of symbols representing the network architecture. The generation of a neural architecture was formulated as the controller's action, whose space is identical to the architecture search space. An alternative to reinforcement learning is evolutionary algorithms, that evolved the topology of architectures by mutating the best architectures found so far \cite{real2018regularized,xie2017genetic,saxena2016convolutional}. There also exists surrogate model based search methods \cite{liu2017progressive} that utilized sequential model-based optimization as a technique for parameter optimization. However, all the above methods require massive computation during the search, particularly thousands of GPU days. Recent efforts such as \cite{liu2018darts, luo2018neural, pham2018efficient}, utilized several techniques trying to reduce the search cost. \cite{liu2018darts} introduced a real-valued architecture parameter which was jointly trained with weight parameters. \cite{luo2018neural} embedded architectures into a latent space and performed optimization before decoding. \cite{pham2018efficient} utilized architectural sharing among sampled models instead of training each of them individually. Several methods attempt to automatically search architectures with fast inference speed, either explicitly took the inference latency as a constrain \cite{tan2018mnasnet}, or implicitly encoded the topology information \cite{guo2018irlas}. Note that \cite{cubuk2018autoaugment} utilized a similar controller inspired by \cite{zoph2017learning}, whose training is time consuming, to guide the augmentation policy search. Our auto-augmentation strategy is much more efficient and economical compared to these methods.

\section{Approach}

In this section, we first present the formulation of hyper-parameter optimization for auto-augmentation strategy. Then we introduce the framework of solving the optimization in an online manner. Finally we detail the search space and training pipeline. The pipeline of our OHL-Auto-Aug is shown in Figure \ref{fig:framework}.

\subsection{Problem Formulation}

The purpose of auto-augmentation strategy is to automatically
find a set of augmentation operations performed
on training data to improve the model generalization according
to the dataset. In this paper, we formulate the augmentation
strategy as $p_{\theta}$, a probability distribution on augmentation operations. Suppose we have $K$ candidate data augmentation operations ${\{O_k(\cdot)\}}_{k=1:K}$, each of which is selected under a probability of $p_{\theta}(O_k)$.
Given a network model $\mathcal{F}(\cdot,w)$ parameterized by $w$, train dataset $\mathcal{X}_{tr}=\{(x_i,y_i)\}_{i=1}^{N_tr}$, and validation dataset $\mathcal{X}_{val}=\{(\hat{x_i},\hat{y_i})\}_{i=1}^{N_val}$, the purpose of data augmentation is to maximum the validation accuracy with respect to $\theta$, and the weights $w$ associated with the model are obtained by minimizing the training loss:
\begin{align}\label{object}
& \max_{\theta} \mathcal{J}(\theta) = acc(w^*) \\
{\rm{s.t.}} \quad &{w^*}  = \mathop{\arg}\mathop{\min} \limits_{w} \frac{1}{N_{tr}}\sum_{(x,y)\in {\mathcal{X}_{tr}}}\mathbb{E}_{p_{\theta}(O)}[ {\mathcal{L}}(\mathcal{F}(O(x),w),y)], \nonumber
\end{align}
where $acc(w^*) = \frac{1}{N_{tr}}\sum_{(\hat{x},\hat{y})\in {\mathcal{X}_{val}}} \delta(\mathcal{F}(\hat{x},w^*),\hat{y})$, $\mathcal{L}$ denotes the loss function and the input training data is augmented by the selected candidate operations.

This refers to a bilevel optimization problem \cite{colson2007overview}, where augmentation distribution parameters $\theta$ are regarded as hyper-parameters. At the outer level we look for a augmentation distribution parameter $\theta$, under which we can obtain the best performed model $\mathcal{F}(\cdot,{w^*})$, where $w^*$ is the solution of the inner level problem. Previous state-of-theart
methods resort to sampling augmentation strategies with
a surrogate model, and then solve the inner optimization
problem exactly for each sampled strategy, which raises the
time-consuming issue. To tackle this problem, we propose
to train the augmentation distribution parameters alongside
with the training of network weights, getting rid of training
thousands of networks from scratch.

\subsection{Online Optimization framework}

Inspired by the forward hyper-gradient proposed in \cite{franceschi2017forward}, we propose an online optimization framework that trains the hyper-parameters and the network weights simultaneously. To clarify, we use $T=1,2,...T_{max}$ to denote the update
steps of the outer loop, and $i = 1,2,...,I$ to denote the update
iterations of the inner loop. This indicates that between
two adjacent outer optimization updates, the inner loop updates
for $I$ steps. Each steps train the network weights with
a batch size of $B$ images. In this section, we focus on the
process of the $T$-th period of the outer loop.

For each image, a augmentation operation is sampled
according to the current augmentation policy distribution
$p_{\theta_{T-1}}$. We use trajectory $\mathcal{T}=\{O_{k(j)}\}_{j=1:(I\times B)}$ to denote
all the augmentation operations sampled in the $T$-th period. For the $I$ iterations of the inner loop, we have
\begin{align}\label{update_inner}
&w_{T-1}^{(i)} = w_{T-1}^{(i-1)} - \eta_w \nabla_{w}[\mathcal{L}_B(\mathcal{T}^{(i)},w_T^{(i-1)},x_B,y_B)], \nonumber\\
&w_{T-1}^{(0)} = w_{T-1}, w_{T} = w_{T-1}^{(I)},
\end{align}
where $\eta_w$ is learning rate for the network parameters, $\mathcal{L}_B$ denotes the batched loss which takes three inputs: the augmentation operations $\mathcal{T}^{(i)}$ for the $i$-th batch, the current inner loop parameters $w_T^{(i-1)}$ and a mini-batched data $x_B$, $y_B$, and returns an average loss on the $i$-th batch.

As illustrated in Equation \ref{update_inner}, $w_T$ is affect by the trajectory $\mathcal{T}$. Since $w_T$ is only influenced by the operations
in trajectory $\mathcal{T}$, $w_T$ should be regarded as a random variable,
with probability $p(w_T)=p(\mathcal{T})$, where $p(w_T)$ denotes
the probability of $\mathcal{T}$ sampled by $p_{\theta_{T-1}}$, calculated as $p(\mathcal{T})=\prod_{j=1}^{I\times B}p_{\theta_{T-1}}(O_{k(j)})$. We update the augmentation
distribution parameters $\theta_{T}$ by taking one step of optimization
for the outer objective,
\begin{eqnarray}\label{update_outer}
\begin{aligned}
\theta_{T} & = \theta_{T-1} + \eta_{\theta}\nabla_{\theta}\mathcal{J}(\theta_{T-1})\\
& = \theta_{T-1} + \eta_{\theta}\nabla_{\theta} \mathbb{E}_{w_T}[acc(w_T)],
\end{aligned}
\end{eqnarray}
where $\eta_{\theta}$ is learning rate for the augmentation distribution
parameters. As the outer objective is a maximization, here
the update is ‘gradient ascent’ instead of ‘gradient descent’.
The above process iteratively continues until the network
model is converged.

The motivation of this online framework is that we approximate $w^*(\theta)$ using only $I$ steps of training, instead of completely solving the inner optimization in Equation \ref{object} until convergence. 
At the outer level, we would like to find
the augmentation distribution that can train a network to
a high validation accuracy given the $I$-step optimized network
weights. A similar approach has been exploited in
meta-learning for neural architecture search \cite{liu2018darts}. Although
the convergence are not guaranteed, practically the optimization
is able to reach a fixed point as our experiments
have demonstrated, as well in \cite{liu2018darts}.

It is a tricky problem to calculate the gradient of validation accuracy with respect to $\theta$ in Equation \ref{update_outer}. 
This is mainly due to two facts: 1. the validation accuracy is non-differentiable
with respect to $\theta$. 2. Analytically calculating the integral over $w_T$ is intractable. To address these
two issues, we utilize REINFORCE \cite{williams1992simple} to approximate
the gradient in Equation \ref{update_outer} by Monte-Carlo sampling. This
is achieved by training $N$ networks in parallel for the inner
loop, which are treated as $N$ sampled trajectories. We
calculate the average gradient of them based on the REINFORCE
algorithm. We have
\begin{align}\label{Reinforce}
\nabla_{\theta}\mathcal{J}(\theta)& = \nabla_{\theta}\mathbb{E}_{w_T}[acc(w_T)]
\approx\frac{1}{N}\sum_{n=1}^{N}\nabla_{\theta} log(p(w_{T,n}))acc(w_{T,n}) \nonumber\\
& = \frac{1}{N}\sum_{n=1}^{N}\nabla_{\theta} log(p(\mathcal{T}_n))acc(w_{T,n}), 
\end{align}
where $\mathcal{T}_n$ is the $n$-th trajectory. By substituting $p(\mathcal{T}_n)$ to Equation \ref{Reinforce},
\begin{eqnarray}\label{Reinforce_2}
\begin{aligned}
\nabla_{\theta}\mathcal{J}(\theta)&\approx \frac{1}{N}\sum_{n=1}^{N}\nabla_{\theta} log(p(\mathcal{T}_n))acc(w_{T,n})\\
&=\frac{1}{N}\sum_{n=1}^{N}\sum_{j=1}^{I\times B}\nabla_{\theta} log(p_{\theta_{T-1}}(O_{k(j),n}))acc(w_{T,n}).
\end{aligned}
\end{eqnarray}

In practice, we also utilize the baseline trick \cite{silver2016mastering} to reduce
the variance of the gradient estimation, 
\begin{eqnarray}\label{Reinforce_baseline}
\begin{aligned}
\nabla_{\theta}\mathcal{J}(\theta) \approx \frac{1}{N}\sum_{n=1}^{N}\sum_{j=1}^{I\times B}\nabla_{\theta} log(p_{\theta_{T-1}}(O_{k(j),n}))\tilde{acc}(w_{T,n}),
\end{aligned}
\end{eqnarray}
where $\tilde{acc}(w_{T,n})$ denotes the function that normalizes the
accuracy returned by each sample trajectory to zero mean. \footnote{To meet $p(w_T)=p(\mathcal{T})$, we use the same sampled mini-batches for all trajectories, so there is no randomness caused by mini-batch sampling.}

Since we have $N$ trajectories, each of which outputs a
$w_{T,n}$ from Equation \ref{update_inner}, we need to synchronize these $w_{T,n}$
to get a same start point for the next training step. We simply
select the $w_{T,n}$ with the best validation accuracy as the
final output of Equation \ref{update_inner}.

\subsection{Search Space and Training Pipeline}

In this paper, we formulate the auto-augmentation strategy
as distribution optimization. For each input train image,
we sample an augmentation operation from the search
space and apply to it. Each augmentation operation consists
of two augmentation elements. The candidate augmentation
elements are listed in Table \ref{list-aug}.

\begin{table}
\newcommand{\tabincell}[2]{\begin{tabular}{@{}#1@{}}#2\end{tabular}}  
\caption{List of Candidate Augmentation Elements}\label{list-aug}
\begin{tabular}{c|c}
\hline

 {Elements Name} & range of magnitude\\
\hline
$Horizontal Shear$ &  $\{0.1, 0.2, 0.3\}$ \\
$Vertical Shear$   &  $\{0.1, 0.2, 0.3\}$\\
$Horizontal Translate$& $\{0.15, 0.3, 0.45\}$\\
$Vertical Translate$ &$\{0.15, 0.3, 0.45\}$\\
$Rotate$    &          $\{10, 20, 30\}$\\
$Color Adjust$    &     $\{0.3, 0.6, 0.9\}$\\
$Posterize$ &        $\{4.4, 5.6, 6.8\}$\\
$Solarize$     &     $\{26, 102, 179\}$\\
$Contrast$    &    $\{1.3, 1.6, 1.9\}$\\
$Sharpness$  &     $\{1.3, 1.6, 1.9\}$\\
$Brightness$    &  $\{1.3, 1.6, 1.9\}$\\
$AutoContrast$ & None \\
$Equalize$  & None\\
 $Invert$ &  None\\
\hline
\end{tabular}
\end{table}

The total number of the elements is $36$. We will sample augmentation choices twice,  The augmentation distribution is a multinomial distribution which has $K$ possible outcomes. The probability of the $k$th operation is a normalized sigmoid function of $\theta_k$.
A single operation is defined as the combination of two elements, resulting in $K=36^2$ possible combinations (the same augmentation choice can repeat). The augmentation distribution is a multinomial distribution which has $K$ possible outcomes. The probability of the $k$th operation is a normalized sigmoid function of $\theta_k$,
\begin{eqnarray}
\begin{aligned}
 p_{\theta}(O_k) = \frac{(\frac{1}{1+e^{-\theta_k}})}{\sum_{i=1}^{K}(\frac{1}{1+e^{-\theta_i}})},
\end{aligned}
\end{eqnarray}
where $\theta \in \mathbb{R}^{K}$.


The whole training pipeline is described in Algorithm \ref{alg:1}.

\section{Experiments}
\subsection{Implementation Details} \label{Implement}

\begin{algorithm}[tb]
\caption{\small{Online Optimization for Auto-Augmentation Strategy}}
\label{alg:1}
\begin{algorithmic}
\small{

\STATE {Initialize $\theta_0$, initialize the same $w_0$ for $N$ models;} 
\WHILE {$T \le T_{max}$} 
\FORALL{$n$ such that $1\leq n\leq N$}
\FORALL{$i$ such that $0\leq i\leq I$}
\STATE Compute $\{w_{T,n}^{i}\}$ in Equation \ref{update_inner};
\ENDFOR

return $w_{T,n}$
\ENDFOR

return $\{w_{T,n}\}_{n=1:N}$
\STATE Fix $\{w_{T,n}\}_{n=1:N}$, calulate $\nabla_{\theta}\mathcal{J}(\theta)$ in Equation \ref{Reinforce_baseline};
\STATE Update $\theta_T$ according to Equation \ref{update_outer};
\STATE Select $w_T$ from $\{w_{T,n}\}_{n=1:N}$ with the best validation accuracy;
\STATE Broadcast $w_T$ to all the $N$ models;
\ENDWHILE
\STATE return $w_{T_{max}}$, $\theta_{T_{max}}$;}
\end{algorithmic}
\end{algorithm}

\begin{table*}[h]
\centering
\caption{Test error rates (\%) on CIFAR-10. The number in brackets refers to the results of our implementation. We compare our OHL-Auto-Aug with standard augmentation (Baseline), standard augmentation with Cutout (Cutout), augmentation strategy discovered by \cite{cubuk2018autoaugment} (Auto-Augment). Compared to Baseline, our OHL-Auto-Aug achieves about 30\% reduction of error rate. }
\newcommand{\tabincell}[2]{\begin{tabular}{@{}#1@{}}#2\end{tabular}}  
\label{tb:CIFAR}
\small
\begin{tabular}{c | c | c | c | c | c}
    \hline
    {Model} & Baseline & Cutout \cite{devries2017improved} & Auto-Augment \cite{cubuk2018autoaugment}& OHL-Auto-Aug & \tabincell{c}{Error Reduce\\(Baseline/Cutout)} \\
    \hline
    \hline
    ResNet-18 \cite{he2016deep}                                  & 4.66 & 3.62 & 3.46 & \textbf{3.29} & \textbf{1.37/0.33}\\
    \hline
    WideResNet-28-10 \cite{zagoruyko2016wide}                    & 3.87 & 3.08 & 2.68 & \textbf{2.61} & \textbf{1.26/0.47}\\
    \hline
    DualPathNet-92 \cite{chen2017dual}                           & 4.55 & 3.71 & 3.16 & \textbf{2.75} & \textbf{1.8/0.96}\\
    \hline
    \tabincell{c}{AmoebaNet-B(6, 128) \cite{cubuk2018autoaugment} \\ (our impl.)}  & \tabincell{c}{2.98 \\ (3.4)} & \tabincell{c}{2.13 \\ (2.9)} & \textbf{1.75} & 1.89 & \textbf{\tabincell{c}{1.09/0.24 \\ (1.51/1.01)}}\\
    \hline
\end{tabular}
\end{table*}

We perform our OHL-Auto-Aug on two classification datasets: CIFAR-10 and ImageNet. Here we describe the experimental details that we used to search the augmentation policies.

\textbf{CIFAR-10} CIFAR-10 dataset \cite{krizhevsky2014cifar} consists of natural images with resolution 32$\times$32. There are totally 60,000 images in 10 classes, with 6000 images per class. The train and test sets contain 50,000 and 10,000 images respectively. For our OHL-Auto-Aug on CIFAR-10, we use a validation set of 5,000 images, which is randomly splitted from the training set, to calculate the validation accuracy during the training for the augmentation distribution parameters. 

In the training phase, the basic pre-processing follows the convention for state-of-the-art CIFAR-10 models: standardizing the data, random horizontal flips with 50\% probability, zero-padding and random crops, and finally
Cutout \cite{devries2017improved} with 16$\times$16 pixels. Our OHL-Auto-Aug strategy is applied in addition to the basic pre-processing. For each training input data, we first perfom basic pre-processing, then our OHL-Auto-Aug strategy, and finally Cutout.

All the networks are trained from scartch on CIFAR-10. For the training of network parameters, we use a mini-batch size of 256 and a standard SGD optimizer. The momentum rate is set to 0.9 and weight decay is set to 0.0005. The cosine learning rate scheme is utilized with the initial learning rate of 0.2. The total training epochs is set to 300. For the AmoebaNet-B \cite{real2018regularized}, there are several hyper-parameters modifications. The initial learning rate is set to 0.024. We also use an additional learning rate warmup stage \cite{goyal2017accurate} before training. The learning rate is linearly increased from 0.005 to the initial learning rate 0.024 in 40 epochs. 

\textbf{ImageNet} ImageNet dataset \cite{deng2009imagenet} contains 1.28 million training images and 50,000 validation images from 1000 classes. For our experiments on CIFAR-10, we set aside an additional validation set of 50,000 images splitted from the training dataset.

For basic data pre-processing, we follow the standard practice and perform the random size crop to 224$\times$224 and random horizontal flipping. The practical mean channel substraction is adopted to normalize the input images for both training and testing. Our OHL-Auto-Aug operations are performed following this basic data pre-processing. 

For the training on ImageNet, we use synchronous SGD with a Nestrov momentum of 0.9 and 1e-4 weight decay. The mini-batch size is set to 2048 and the base learning rate is set to 0.8. The cosine learning rate scheme is utilized with a warmup stage of 2 epochs. The total training epoch is 150. 

\textbf{Augmentation Distribution Parameters}
For the training of augmentation policy parameters, we use Adam optimizer with a learning rate of $\eta_{\theta}=0.05$, $\beta_1=0.5$ and $\beta_2=0.999$. On CIFAR-10 dataset, the number of trajectory sample is set to 8. This number is reduced to 4 for ImageNet, due to the large computation cost. Having finished the outer update for the distribution parameters, we broadcast the network parameters with the best validation accuracy to other sampled trajectories using multiprocess utensils.

\subsection{Results}

\textbf{CIFAR-10 Results:} 
On CIFAR-10, we perform our OHL-Auto-Aug on three popular network architectures: ResNet \cite{he2016deep}, WideResNet \cite{zagoruyko2016wide}, DualPathNet \cite{chen2017dual} and AmoebaNet-B \cite{real2018regularized}. For ResNet, a 18-layer ResNet with 11.17M parameters is used. For WideResNet, we use a 28 layer WideResNet with a widening factor of 10 which has 36.48M parameters. For Dual Path Network, we choose DualPathNet-92 with 34.23M parameters for CIFAR-10 classification. AmoebaNet-B is an auto-searched architecture by a regularized evolution approach proposed in \cite{real2018regularized}. We use the AmoebaNet-B (6,128) setting for fair comparison with \cite{cubuk2018autoaugment}, whose number of parameters is 33.4M. The WideReNet-28, DualPathNet-92 and AmoebaNet-B are relatively heavy architectures for CIFAR-10 classification. 

We illustrate the test set error rates of our OHL-Auto-Aug for different network architectures in Table \ref{tb:CIFAR}. All the experiments follow the same setting described in Section \ref{Implement}. We compare the empirical results with standard augmentation (Baseline), standard augmentation with Cutout (Cutout), an augmentation strategy proposed in \cite{cubuk2018autoaugment} (Auto-Augment). For ResNet-18 and DualPathNet-92, our OHL-Auto-Aug achieves a significant improvement than Cutout and Auto-Augment. Compare to the Baseline model, the error rate is reduced about $30\%$ ResNet-18, about $40\%$ for DualPathNet-92 and about $30\%$ for WideResNet-28-10. For AmoebaNet-B, although we have not reproduced the accuracy reported in \cite{cubuk2018autoaugment}, our OHL-Auto-Aug still achieves a 1.09\% error rate redunction compared to Baseline model. Compared to the baseline we implemented, our OHL-Auto-Aug achieves an error rate of $1.90\%$ which is about $45\%$ error rate reduction. For all the models, our OHL-Auto-Aug exhibits a significantly error rate drop compared with Cutout.

We also find that the results of our OHL-Auto-Aug outperform those of Auto-Augment on all the networks except AmoebaNet-B. We think the main reason is that we could not train the Baseline AmoebaNet-B to a comparable accuracy with the original paper. By applying our OHL-Auto-Aug, the accuracy gap can be narrowed from 0.42\% (Baseline accuracy comparing \cite{cubuk2018autoaugment} with our implementation) to 0.14\% (comparing OHL-Auto-Aug with Auto-Augment), which could demonstrate the effectiveness of our approach.

\textbf{ImageNet Results:} 
On ImageNet dataset, we perform our method on two networks: ResNet-50 \cite{he2016deep} and SE-ResNeXt101 \cite{hu2018squeeze} to show the effectiveness of our augmentation strategy search. ResNet-50 contains 25.58M parameters and SE-ResNeXt101 contains 48.96M parameters. We regard ResNet-50 as a medium architecture and SE-ResNeXt101 as a heavy architecture to illustrate the generalization of our OHL-Auto-Aug.

\begin{table}[h]
\centering
\caption{Top-1 and Top-5 error rates (\%) on ImageNet. We compare our OHL-Auto-Aug with standard augmentation (Baseline), standard augmentation with mixup \cite{zhang2017mixup} (mixup), augmentation strategy discovered by \cite{cubuk2018autoaugment} (Auto-Augment). For both the ResNet-50 and SE-ResNeXt-101, our OHL-Auto-Aug improves the performance significantly.}
\newcommand{\tabincell}[2]{\begin{tabular}{@{}#1@{}}#2\end{tabular}}  
\label{tb:ImageNet}
\small
\begin{tabular}{c | c | c  }
    \hline
    {Method} & ResNet-50 \cite{he2016deep}  & SE-ResNeXt-101 \cite{hu2018squeeze} \\ 
    \hline
    \hline
    Baseline                                 &  24.70/7.8  & 20.70/5.01 \\
    \hline
    mixup $\alpha=0.2$ \cite{zhang2017mixup} &   23.3/6.6  & -- \\
    \hline
    Auto-Augment \cite{cubuk2018autoaugment} & 22.37/6.18  & 20.03/5.23 (our impl.) \\
    \hline
    \hline
    OHL-Auto-Aug                             & \textbf{21.07/5.68} & \textbf{19.30/4.57}\\
    \hline
\end{tabular}
\end{table}

\begin{table*}[h]
\centering
\newcommand{\tabincell}[2]{\begin{tabular}{@{}#1@{}}#2\end{tabular}}  
\label{tb:Cost}
\caption{Search cost of our method compared with Auto-Augment \cite{cubuk2018autoaugment}. For fair comparison, we compute the total training iterations with conversion under a same batch size 1024 and denote as `$\#Iterations$'. Specific computing implementation is detailed in Section \ref{SearchCost}. Our OHL-Auto-Aug achieves 60$\times$ faster on CIFAR-10 and 24$\times$ faster on ImageNet than Auto-Augment. Our OHL-Auto-Aug also gets rid of the need of retraining from scratch, further saving computation resources.}
\label{tb:SearchCost}
\small
\begin{tabular}{c | c | c | c | c }
    \hline
    \multirow{2}*{Dataset}  & \multicolumn{2}{c}{Auto-Augment \cite{cubuk2018autoaugment}} & \multicolumn{2}{|c}{OHL-Auto-Aug} \\
    \cline{2-5}
    ~ & $\# Iterations$& \tabincell{c}{Usage of \\ Dataset (\%)} & $\# Iterations$ &\tabincell{c}{Usage of \\Dataset (\%)} \\
    \hline
    \hline
    CIFAR-10          & $7.03\times10^6$ &$8\%$       & $1.17\times10^5$ &$100\%$ \\
    ImageNet          & $1.76\times10^7$ &$0.5\%$   & $7.5\times10^5$ &$100\%$ \\
    \hline
    \hline
    No Need to Retrain &      \multicolumn{2}{c}{$\times$}   & \multicolumn{2}{|c}{\Checkmark}\\
    \hline
\end{tabular}
\end{table*}

In Table \ref{tb:ImageNet}, we show the Top-1 and Top-5 error rates on validation set. We notice there exists another augmentation stategy mixup \cite{zhang2017mixup} which trains network on convex combinations of pairs of examples and their labels and achieves performance improvement on ImageNet dataset. We report the results of mixup together with Auto-Augment for comparision. All the experiments are conducted following the same setting described in Section \ref{Implement}. Since the original paper of \cite{cubuk2018autoaugment} does not perform experiments on SE-ResNeXt101, we train it with the searched augmentation policy provided in \cite{cubuk2018autoaugment}. As can be seen from the table, OHL-Auto-Aug achieves state-of-the-art accuracy on both models. For the medium model ResNet-50, our OHL-Auto-Aug increases by 3.7\% points over the Baseline, 1.37\% points over Auto-Augment, which is a significant improvement. For the heavy model SE-ResNeXt101, our OHL-Auto-Aug still boosts the performance by 1.4\% even with a very high baseline. Both of the experiments demonstrate the effectiveness of our OHL-Auto-Aug.

\begin{figure}[tb]
	\subfigure[]{\label{Analysis_1}
		\includegraphics[width=0.48\linewidth]{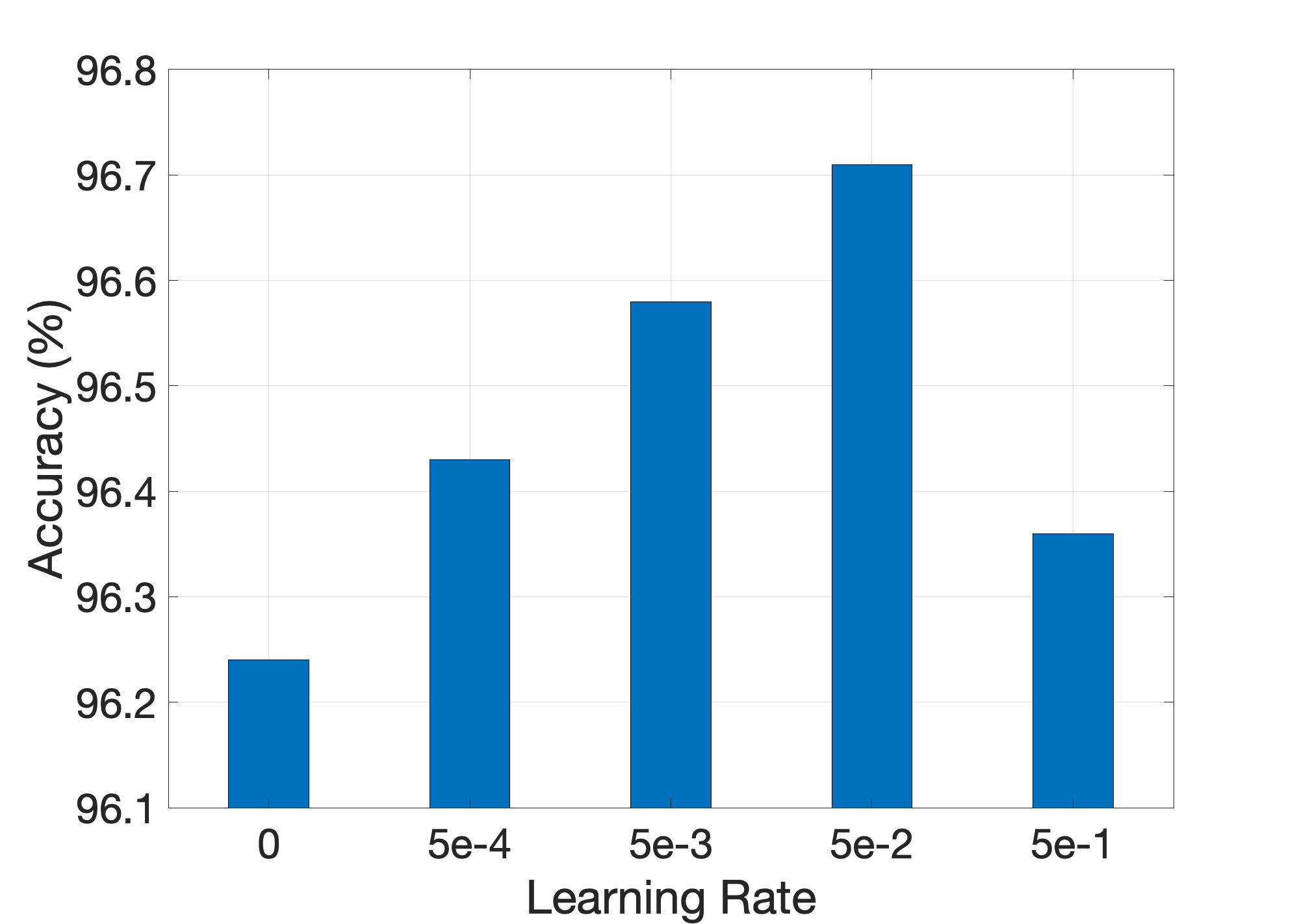}}
	\subfigure[]{\label{Analysis_2}
		\includegraphics[width=0.48\linewidth]{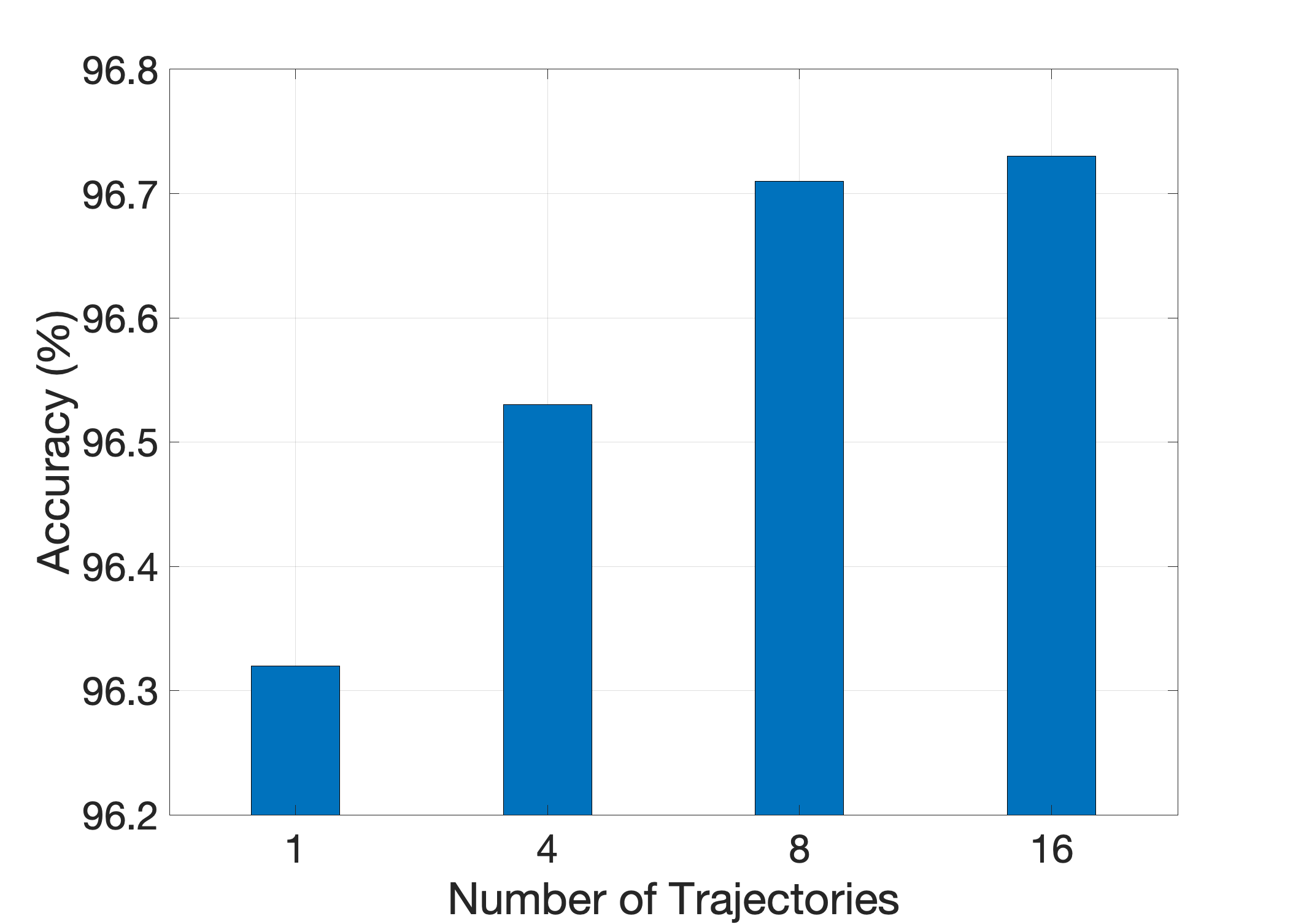}}	
	\caption{\small{(a) Comparisons with different augmentation distribution learning rates. All the experiments are conducted with ResNet-18 on CIFAR-10 following the same setting as described in Section \ref{Implement} except $\eta_{\theta}$}. As can be observed, too large $\eta_{\theta}$ will make the distribution parameters difficult to converge, while too small $\eta_{\theta}$ will slow down the convergency process, both harming the final performance. The chosen of $\eta_{\theta}=5\times10^{-2}$ is a trade-off between convergence speed and performance. (b) Comparisons with different number of trajectory samples.  All the experiments use ResNet-18 on CIFAR-10 following the same setting as described in Section \ref{Implement} except $N$. As can be obseved, increasing the number of trajectories from 1 to 8 steadily improves the performance of the network. When the number comes to $\ge8$, the accuracy improvement is minor. We select $N=8$ on CIFAR-10 as a trade-off between computation cost and performance.}
\vspace{-14pt}
\end{figure}

\textbf{Search Cost:} \label{SearchCost}
The principle motivation of our OHL-Auto-Aug is to improve the efficiency of the search of auto-augmentation strategy. To demonstrate the efficiency of our OHL-Auto-Aug, we illustrate the search cost of our method together with Auto-Augment in Table \ref{tb:SearchCost}. For fair comparison, we compute the total training iterations with conversion under a same batch size 1024 and denote as `$\#Iterations$'. For example, Auto-Augment perfoms searching on CIFAR-10 by sampling 15,000 child models, each trained on a `reduced CIFAR-10' with 4,000 randomly chosen images for 120 epochs. So the $\#Iterations$ of Auto-Augment on CIFAR-10 could be calulated as $15,000\times4,000\times 120 / 1024 = 7.03\times10^6$. Similarly, Auto-Augment samples $15,000$ child models and trains them each on a `reduced ImageNet' of 6,000 images for 200 epochs, which equals to $1.76\times10^7$ $\#Iterations$. Our OHL-Auto-Aug trains on the whole CIFAR-10 for 300 epochs with 8 trajectory samples, which equals to $1.17\times10^5$ $\#Iterations$. For ImageNet, the $\#Iterations$ is $1.28\times10^6\times4\times150/ 1024=7.5\times10^5$.

As illustrated in Table \ref{tb:SearchCost}, even training on the whole dataset, our OHL-Auto-Aug achieves 60$\times$ faster on CIFAR-10 and 24$\times$ faster on ImageNet than Auto-Augment. Instead of training on a reduced dataset, the high efficiency enables our OHL-Auto-Aug to train on the whole dataset, a guarantee for better performance. This also ensures our OHL-Auto-Aug algorithm could be easily performed on large dataset such as ImageNet without any bias on input data. Our OHL-Auto-Aug also gets rid of the need of retraining from scratch, further saving computation resources.

\subsection{Analysis}
In this section, we perform several analysis to illustrate the effectiveness of our proposed OHL-Auto-Aug.


\textbf{Comparisons with different augmentation distribution learning rates:}
We conduct our OHL-Auto-Aug with different augmentation distribution learning rates: $\eta_{\theta}=0, 5\times10^{-4}, \times10^{-3}, 5\times10^{-2}, 5\times10^{-1}$. All the experiments use ResNet-18 on CIFAR-10 following the same setting as described in Section \ref{Implement} except $\eta_{\theta}$. 
The final accuracies are illustrated in Figure \ref{Analysis_1}. Note that our OHL-Auto-Aug chooses $\eta_{\theta}=5\times10^{-2}$.
Comparing the result of $\eta_{\theta}=0$ with other results, we find that updating the augmentation distribution could produce better performed network than uniformly sampling from the whole search space. As can be observed, too large $\eta_{\theta}$ will make the distribution parameters difficult to converge, while too small $\eta_{\theta}$ will slow down the convergency process. The chosen of $\eta_{\theta}=5\times10^{-2}$ is a trade-off for convergence speed and performance.

\textbf{Comparisons with different number of trajectory samples:}
We further conduct an analysis on the sensetivity of the number of trajectory samples. We produce experiments with ResNet-18 on CIFAR-10 with different number of trajectories: $N=1, 4, 8, 16$. All the experiments follows the same setting as described in Section \ref{Implement} except $N$. The final accuracies are illustrated in Figure \ref{Analysis_2}. 
Principly, a larger number of trajectories will benefit the training for the augmentation distribution parameters due to the more accurate gradient calculated by Equation \ref{Reinforce}. Meanwhile, larger number of trajectories increases the memory requisition and the computation cost. As obseved in Figure \ref{Analysis_2}, increasing the number of trajectories from 1 to 8 steadily improves the performance of the network. When the number comes to $\ge8$, the accuracy improvement is minor. We select $N=8$ as a trade-off between computation cost and performance.

\begin{figure}[tb]
  \subfigure[Visualization of augmentation distribution parameters of ResNet-18 training on CIFAR-10.]{\label{weights_cifar}
    \includegraphics[width=1.0\linewidth]{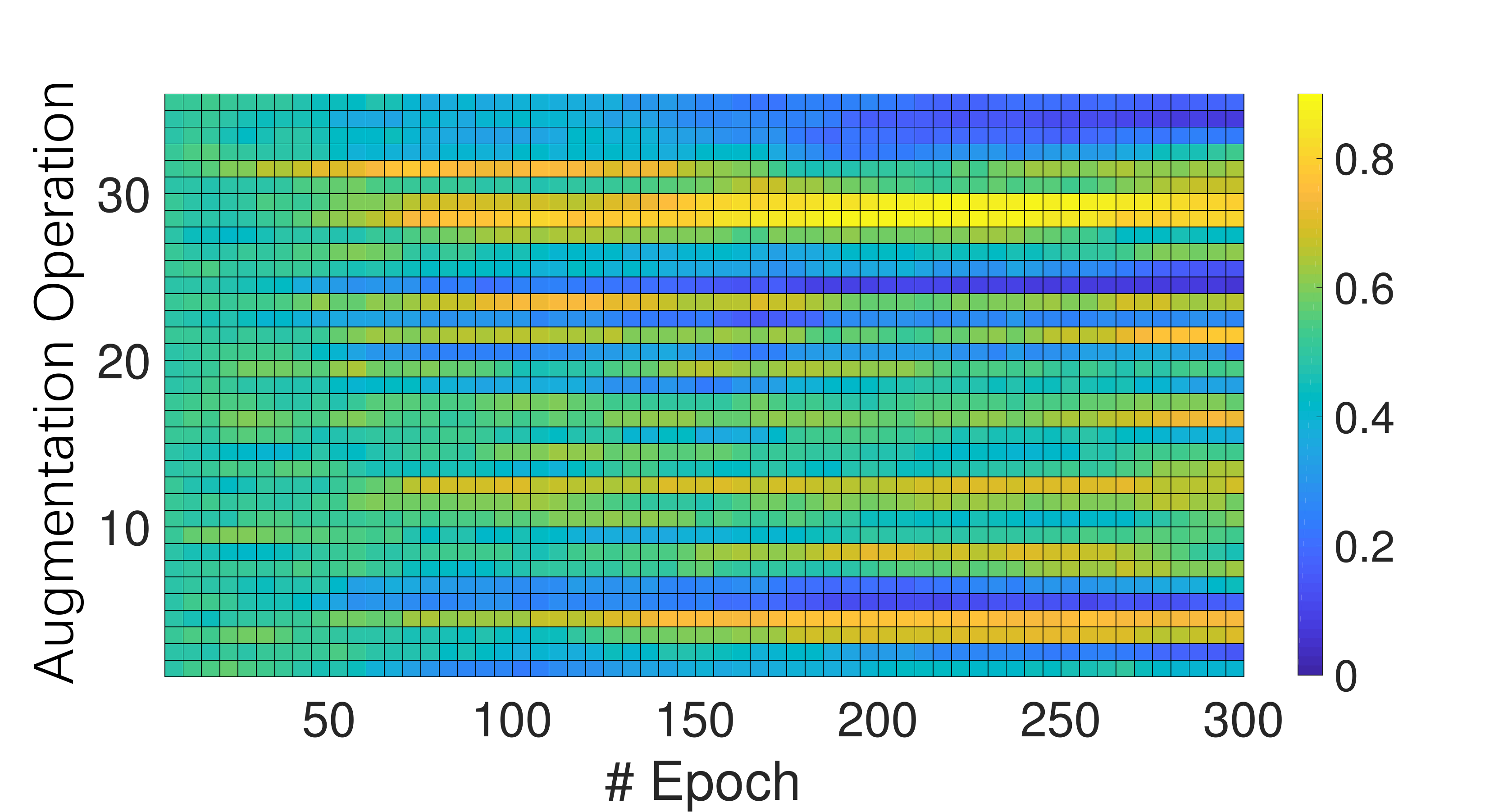}}
  \subfigure[Visualization of augmentation distribution parameters of ResNet-50 training on ImageNet.]{\label{weights_imagenet}
    \includegraphics[width=1.0\linewidth]{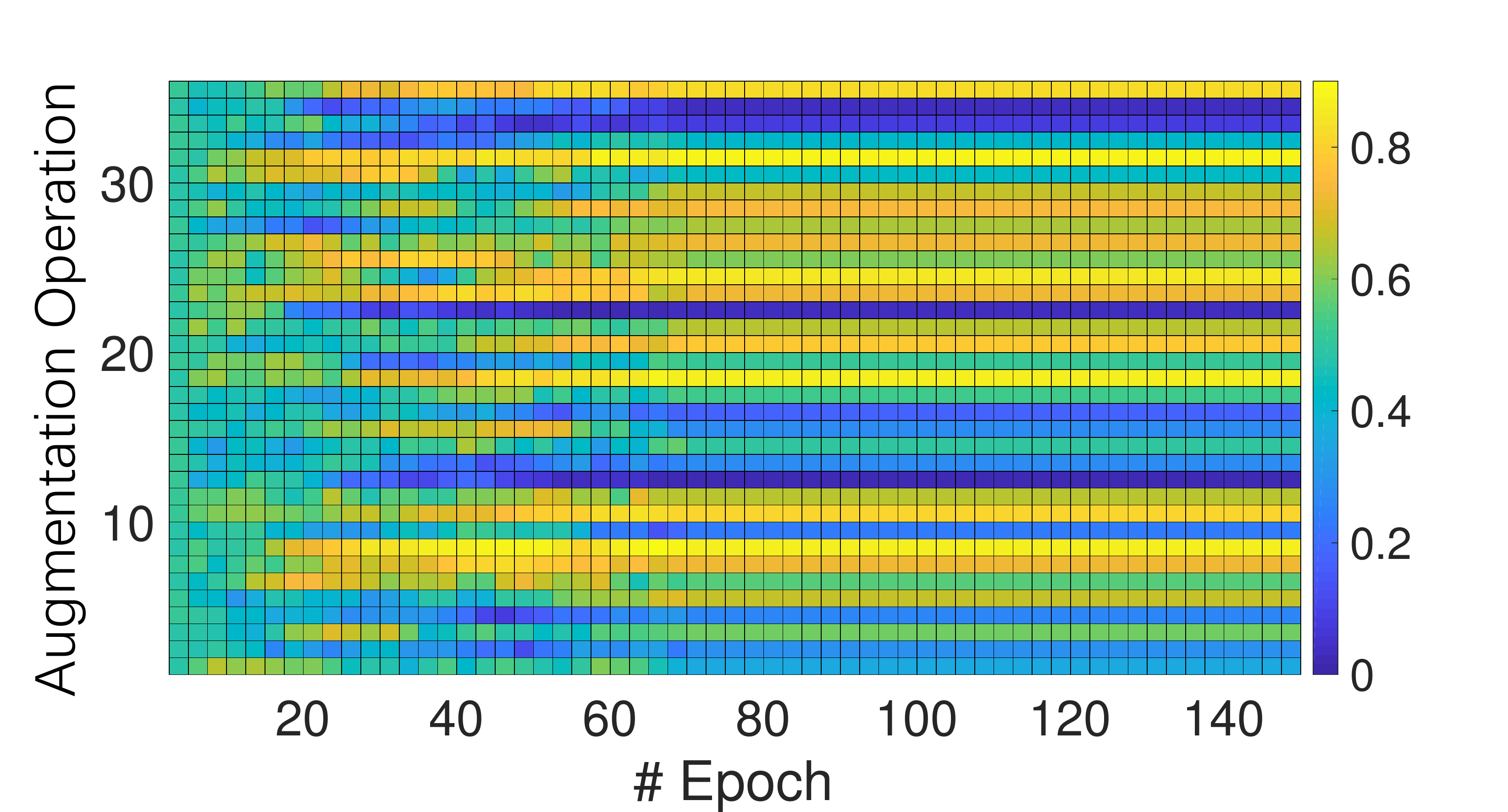}}
  \caption{\small{Analysis of augmentation distribution parameters. For both settings, we find that during training, our augmentation distribution parameters has converged. Some augmentation operations have beed discarded, while the probabilities of some others have increased.}}
  \label{fig-motivation}
\end{figure}

\textbf{Analysis of augmentation distribution parameters:} 
We visualize the augmentation distribution parameters of the whole training stage. We calculate the marginal distribution parameters of the first element in our augmentation operation. Specifically, we convert $\theta$ to a $36\times36$ matrix and sum up each row of the matrix following normalization. The results are illustrated in Figure \ref{weights_cifar} and Figure \ref{weights_imagenet} for CIFAR-10 and ImageNet respectively. On CIFAR-10, we choose the network
of ResNet-18 and show the results for every 5 epochs. On
ImageNet we choose ResNet-50 and show the results for
every 2 epochs. For both the two settings, we find that
during training, our augmentation distribution parameters
has converged. Some augmentation operations have been
discarded, while the probabilities of some others have increased. We also notice that for different dataset, the distributions
of augmentation operations are different. For instance, on CIFAR-10, the 36th operation, which is the `Color Invert', is discarded, while on ImageNet, this operation is preserved with high probability. This further demonstrates our OHL-Auto-Aug could be easily 
performed on different datasets.

\section{Conclusion}
In this paper, we have proposed and online hyper-parameter learning method for auto-augmentation strategy, which formulates the augmentation policy as a parameterized probability distribution. Benefitting from the proposed bilevel framework, our OHL-Auto-Aug is able to optimize the distribution parameters iteratively with the network parameters in an online manner. Experimental results illustrate that our proposed OHL-Auto-Aug achieves remarkable auto-augmentation search efficiency, while establishes significantly accuracy improvements over baseline models. 

\newpage
{\small
\bibliographystyle{ieee_fullname}
\bibliography{egbib}
}

\end{document}